\documentclass[journal]{IEEEtran}

\usepackage{graphicx}
\usepackage{amsmath}
\usepackage{amssymb}
\usepackage{amsmath}
\usepackage{booktabs}
\usepackage{algorithm}
\usepackage{algorithmic}
\usepackage{multirow} 
\usepackage{xfrac}
\usepackage{makecell}
\usepackage[pagebackref,breaklinks,colorlinks]{hyperref}

\usepackage{subcaption}
\usepackage[capitalize]{cleveref}
\crefname{section}{Sec.}{Secs.}
\Crefname{section}{Section}{Sections}
\Crefname{table}{Table}{Tables}
\crefname{table}{Tab.}{Tabs.}

\def\cvprPaperID{16166} 
\def\confName{CVPR}
\def\confYear{2025}

\usepackage{cite}

\usepackage[accsupp]{axessibility}  %
\usepackage{hyperref}

\usepackage{selectp}
\usepackage{orcidlink}

\begin{document}

\title{Hyperbolic Contrastive Learning for Hierarchical 3D Point Cloud Embedding} 
\author{Yingjie Liu, Pengyu Zhang, Ziyao He, Mingsong Chen, Xuan Tang, Xian Wei$^*$\\
East China Normal University\\
{\tt\small $^*$xwei@sei.ecnu.edu.cn}
}
\maketitle

\begin{abstract}
Hyperbolic spaces allow for more efficient modeling of complex, hierarchical structures, which is particularly beneficial in tasks involving multi-modal data.
Although hyperbolic geometries have been proven effective for language-image pre-training, their capabilities to unify language, image, and 3D Point Cloud modalities are under-explored.
We extend the 3D Point Cloud modality in hyperbolic multi-modal contrastive pre-training. Additionally, we explore the entailment, modality gap, and alignment regularizers for learning hierarchical 3D embeddings and facilitating the transfer of knowledge from both Text and Image modalities. These regularizers enable the learning of intra-modal hierarchy within each modality and inter-modal hierarchy across text, 2D images, and 3D Point Clouds.
Experimental results demonstrate that our proposed training strategy yields an outstanding 3D Point Cloud encoder, and the obtained 3D Point Cloud hierarchical embeddings significantly improve performance on various downstream tasks.
%
\end{abstract}


\section{Introduction}
\label{sec:intro}
Recently, language models (LMs) \cite{nie2024text, yang2024enhancing, mandica2024hyperbolic,behnamghader2024llm2vec} have made great progress and shown remarkable capabilities in understanding and generating natural language.
Meanwhile, to harness the advancements in language models, recent approaches \cite{qi2025shapellm,guo2023point,hong20233d} have evolved to combine visual processing with the reasoning and generalization capabilities of LMs by aligning vision and language embeddings in shared feature space and ensuring consistency.
Even with the significant resource investment and progress in training schemes or prompt engineering, recalling the manifold learning hypothesis reveals that these models still face limitations, particularly due to the lack of consideration for the geometric priors of the feature space.
%
The default Euclidean geometry used in these models for learning embeddings may not always be optimal, particularly in representing complex hierarchical structures and relationships in real-world data, whether in text modality or vision modality.

Hierarchical structure is a fundamental component of the natural world. 
Humans comprehend the world through the relationships and hierarchies described above \cite{rasmussen1985role,palmer1977hierarchical}.
For example, all noun hierarchies lead to an entity, and verb synsets detail events, e.g., ``communicate'' to ``whisper'' \cite{fellbaum2010wordnet}. 
Text-vision pairs also show hierarchy, while pixels form shapes, combining to create scenes, each layer building on the previous to abstract higher.
%
%
Studies have shown that the text and vision data (including 2D images and 3D Point Clouds in this work) are part of the hierarchy \cite{shimizuhyperbolic2021,DBLP:conf/icml/DesaiNR0V23,montanaro2022rethinking}.
Meanwhile, latent embeddings with an underlying tree-like and hierarchical structure learned by deep neural networks exhibit better performance \cite{montanaro2023towards,montanaro2022rethinking,shimizuhyperbolic2021,DBLP:conf/icml/DesaiNR0V23,kim2024hype,mandica2024hyperbolic}.
Most recent research \cite{yang2024enhancing} investigates the non-Euclidean characteristics of LLMs on complex reasoning tasks, finding that token embeddings and hidden states exhibit a significant degree of hyperbolicity, indicating an underlying hyperbolic structure.
We further hypothesize that incorporating hierarchical concepts in model feature space design can help models maintain stable, coherent perception when faced with complex visual inputs, which is beneficial for understanding the real world.
Hyperbolic space with negative curvature is well-suited for modeling hierarchical data, yielding remarkable performance.
Hierarchical embeddings in the hyperbolic space have been previously explored in single-modal and uni-modal settings, learning shared embeddings of different types of modalities, including Text and Images \cite{ge2023hyperbolic,yue2023hyperbolic}.
MERU \cite{DBLP:conf/icml/DesaiNR0V23} is the first large-scale contrastive image-text models that yield hyperbolic embeddings.
\cite{Ramasinghe2024} further considers the modality-gap problem while preserving hierarchies.

However, the aforementioned challenges remain when incorporating LLMs for 3D object understanding in hyperbolic space, especially embodied interaction that relies on precise geometry, which is currently under-explored.
Hence, in this work, we further explore the hierarchical prior in 3D object understanding by bridging the hyperbolic language-image model and 3D Point Cloud representation learning.
We train contrastive text-2D image-3D Point Cloud models that yield hyperbolic embeddings that capture the visual-semantic hierarchy.
We summarize our contributions as follows:
\begin{itemize}
\item We propose a regularizer for point cloud embedding reconstruction to promote \textit{intra-modal} hierarchical knowledge capturing and implement a guidance contrastive learning process, aligning 3D Point Cloud embeddings with hyperbolic Text-Image embeddings.
\item We propose novel hierarchy-enhancing losses that promote \textit{inter-modal} hierarchical concept relations during contrastive learning, achieving hierarchical embeddings for 3D Point Clouds, extending beyond common Text-Image modalities.
\item Experimental results show significant improvements in the performance of various point cloud tasks based on our hierarchical 3D Point Cloud embeddings.

\end{itemize}

\section{Related Work}

\subsection{Hyperbolic Geometry for Point Clouds}

Hyperbolic embedding learning has been explored in various fields \cite{DBLP:journals/corr/NickelK17, DBLP:conf/iclr/CetinCBH23, DBLP:conf/icml/GaneaBH18, DBLP:conf/cvpr/KhrulkovMUOL20, liu2024application}.
Hyperbolic embeddings with InfoNCE loss for predicting hierarchical relations in the WordNet nouns hypernymy tree was first proposed in \cite{DBLP:journals/corr/NickelK17}.
\cite{DBLP:conf/icml/GaneaBH18} suggested entailment loss as an alternative.
In the vision area, \cite{DBLP:conf/cvpr/KhrulkovMUOL20} proposed Hyperbolic ProtoNet for few-shot classification.
\textbf{Point clouds of 3D objects also exhibit an inherent hierarchical compositional nature.}
HyCoRe \cite{montanaro2022rethinking} first proposes the explicit regularization to capture part-whole hierarchies and experimentally observed hierarchical structures, noting that hierarchical structures of embeddings naturally emerge within hyperbolic space but are crude without its proposed regularization \cite{montanaro2023towards}.
PHGT \cite{liu2024application} leverages an attention module based on the Poincaré ball model to enhance 3D Point Cloud feature extraction and classification.
HypLiLoc \cite{wang2023hypliloc} fuses Euclidean and hyperbolic features for improved pose regression.
HECPG \cite{xie2024hecpg} introduces hyperbolic attention with hyperbolic weight and Riemannian metric to fuse hyperbolic features, boosting point cloud matching accuracy adaptively.
We further propose novel pre-training losses that enhance hyperbolicity, thereby preserving hyperbolic modeling capabilities during the multi-modal contrastive learning process.

\subsection{Contrastive Pre-training for Hierarchy Multi-Modal Embeddings}
%
ConVIRT \cite{DBLP:conf/mlhc/0004JMML22} pioneered contrastive pre-training \cite{ilharco_gabriel_2021_5143773,radford2021learning} for zero-shot image classification, maintaining L2-normalization and cosine similarity. 
There are some other variant methods like CoCa \cite{DBLP:journals/tmlr/YuWVYSW22} added captioning loss by a multi-modal text decoder, OTTER \cite{DBLP:conf/iclr/WuCZGGV22} considered intra-modal similarity, and SigLIP \cite{DBLP:conf/iccv/ZhaiM0B23} applied logistic regression.
Other than MERU \cite{DBLP:conf/icml/DesaiNR0V23}, exponentially lifts the embeddings onto the Lorentz hyperboloid, combining entailment learning with the CLIP approach to learn embeddings in hyperbolic space capturing latent visual-semantic hierarchies.
\cite{Ramasinghe2024} further discusses the modality gap in hyperbolic space.
The most recent \cite{pal2024compositional} extends to include image patches and caption parts, enforcing an ordering that reflects the hierarchy shared by both modalities.
EuCLIP \cite{chou2024embedding} captures hierarchical relationships by using Euclidean geometry with negative squared distance softmax logits and removing final layer normalization.
%
%
%

The point cloud is a fundamental modal for understanding the three-dimensional (3D) world.
Both inter and intra-modal contrastive learning strategies are extended to the point cloud area \cite{afham2022crosspoint}.
PointCLIP \cite{zhang2022pointclip, zhu2023pointclip} further aligns point clouds to 2D depth images and text in the context of CLIP.
ReCon \cite{qi2023contrastReCon} utilizes contrast guided by reconstruction to address the pattern disparities between local masked data modeling and global cross-modal alignment. 
ShapeLLM \cite{qi2025shapellm} further scales up the parameters of ReCon and broadens the scale of the pretraining dataset for robust 3D embeddings.
\cite{chen2024grounded} pre-trains a 3D Point Cloud encoder and cross-modal interactor using phrase-level scene-text annotations, then tunes for multi-task instruction with referent tokens for flexible 3D scene understanding.
%
However, it remains under-explored in the context of learning hyperbolic contrastive 3D Point Cloud embeddings. 
We further explore inferring concept hierarchies across multiple modalities, including text, 2D images, and 3D Point Clouds.

\section{Preliminary}

In this section, we introduce relevant notations and briefly review the Lorentz space and related contrastive losses used in this work as preliminary knowledge. 
We divide a dataset of text-vision pairs into mini-batches $\mathcal{B}=\{(T_1, V_1, P_1), (T_2, I_2, P_2), \dots\}$,
where $T_i$, $I_i$ and $P_i$ denotes text, 2D images, and 3D Point Clouds, respectively. 
Let $|\mathcal{B}|$ denote the batch size and $i\in |\mathcal{B}|$.
Further, assume that we have a text encoder $f(\cdot)$, an image encoder $g(\cdot)$, and a 3D Point Cloud encoder $h(\cdot)$.
Let $\mathbf{x}, \mathbf{y}, \mathbf{z}$ denote hyperbolic text embedding, hyperbolic image embedding and hyperbolic point cloud embeddings, respectively.

\subsection{Hyperbolic Geometry \& Lorentz Embeddings}
Hyperbolic space is a non-Euclidean space with constant negative curvature.
%
Following \cite{Ramasinghe2024}, to better avoid numerical instabilities in the training process that comes from exponential volume growth, we adopt the Lorentzian hyperboloid rather than other popular models of hyperbolic geometry, e.g., the Poincaré ball model, the Klein model, the Poincaré half-space model, and the hemisphere model. The Poincaré ball model and Beltrami-Klein model are the projections of the hyperboloid model onto the different dimensional space-like hyperplanes, as more details can referred to in \cite{shimizuhyperbolic2021}.
Taking $\mathbf{x}$ as an example, we provide a background discussion of the hyperbolic space but limit it to the Lorentz / hyperboloid model, denoted by 
\begin{equation*}
    \mathbb{L}^n = \{\mathbf{x} \in \mathbb{R}^{n+1} : \langle \mathbf{x},\mathbf{x} \rangle_\mathbb{L} = \sfrac{-1}{c} \} \; , \; c > 0,
\end{equation*}
where every vector $\mathbf{x}$ can be written as $[\mathbf{x}_{space}, x_{time}]$,  
$x_{time}\in \mathbb{R}$ serves as the axis of symmetry \cite{Ramasinghe2024}. 
Note that for $c > 0$, the curvature is $-c$.
Since $\mathbf{x}$ always lies on the hyperboloid, the time dimension can then be inferred as $x_{time} = \sqrt{\sfrac{1}{c} + \lVert \mathbf{x}_{space} \rVert^2}$.
$\langle \cdot , \cdot \rangle_\mathbb{L}$ denotes the Lorentzian inner product as 
\begin{equation*}
    \langle \mathbf{x},\mathbf{y} \rangle_\mathbb{L} = \mathbf{x}_{space} \cdot \mathbf{y}_{space} - x_{time} \; y_{time}.
\end{equation*}
%
We only consider these maps where $\mathbf{m}$ is the origin of the hyperboloid ($\mathbf{O} = [\mathbf{0}, \sfrac{1}{c}]$) \cite{Ramasinghe2024}, i.e., simplifying the exponential map by using the tangent space of the origin, thus meanwhile minimizing potential numerical instability in the model’s computation \cite{kim2024hype}.
The exponential map provides a way to map vectors from tangent spaces onto the manifold. 
For the point $\mathbf{O}$ on the hyperboloid, it is defined as $\text{expm}_{\mathbf{O}}: \mathcal{T}_\mathbf{O} \mathbb{L}^n\rightarrow \mathbb{L}^n$.
Let $\mathbf{u} = [\mathbf{u}_{enc} , 0] \in \mathbb{R}^{n+1}$, thus $\langle\mathbf{O}, \mathbf{u}\rangle = 0$ and $\mathbf{u}$ belong to the tangent space at the hyperboloid origin $\mathbf{O}$ .
We have the space dimension of hyperbolic text embedding as $\mathbf{x}_{space} = \text{expm}_{\mathbf{O}, space}(\mathbf{u})$, by lifting the embeddings to the Lorentz hyperboloid $\mathbb{L}^n$ through the exponential map
\begin{equation*}
    \text{expm}_{\mathbf{O}, space}(\mathbf{u}) = \frac{\sinh(\sqrt{c} \; \lVert \mathbf{u}_{enc} \rVert)}{\sqrt{c} \; \lVert \mathbf{u}_{enc} \rVert} \mathbf{u}_{enc} + 0.
\end{equation*}
Note that $\mathbf{u}_{enc}$ denote the scaled text embeddings,
\begin{align*} 
\mathbf{u}_{enc} =  \alpha_{txt}f(T).
\end{align*}
Specifically, $f(T)\in \mathbb{R}^n$ would have an expected norm $\sqrt{n}$ and the exponential map scales it to $e^{\sqrt{n}}$, which can be numerically large~\cite{DBLP:conf/icml/DesaiNR0V23}.
Both $\alpha_{txt}$ are initialized to $\frac{1}{\sqrt{n}}$ and learned in logarithmic space to avoid collapsing all embeddings to zero so that the embeddings have an expected unit norm at initialization, preventing numerical overflow.
Similarly, for the image modal, we have $\mathbf{v}_{enc} = \alpha_{img}g(I)\in \mathbb{R}^n$ and further obtain hyperbolic image embedding $\mathbf{y}$.

A geodesic is the shortest path between two points on the manifold. 
Geodesics in the Lorentz model are curves traced by the intersection of the hyperboloid with hyperplanes passing through the origin of $\mathbb{R}^{n+1}$ 
\begin{equation*}
 d_\mathcal{L} (\mathbf{x}, \mathbf{y}) = \sqrt{\sfrac{1}{c}} \cdot \cosh^{-1} ( -c \; \langle \mathbf{x},\mathbf{y} \rangle_\mathbb{L}).
\end{equation*}

\subsection{Language-Vision Contrastive Learning Loss}
\noindent\textbf{Contrastive Loss for Hyperbolic Language-Image Embeddings}~ 
Considering language-image contrastive learning loss $\mathcal{L}_{\textbf{cont}}$, it can formulated by applying InfoNCE~\cite{DBLP:journals/corr/abs-1807-03748} with similarity function that measures the relationship within pairs.
%
The similarity function is cosine similarity in CLIP, i.e., $\mathrm{sim}(\mathbf{x}, \mathbf{y}) = \frac{\mathbf{x} \cdot \mathbf{y}}{\|\mathbf{x}\| \|\mathbf{y}\|}$ where $\| \cdot \|$ is the L2 norm. 
%
For hyperbolic contrastive learning, we follow the formulation and the hyperboloid model in MERU~\cite{DBLP:conf/icml/DesaiNR0V23} whose similarity function is parameterized by three trainable scalars: text embedding scale $\alpha_{txt}$, image embedding scale $\alpha_{img}$, and curvature parameter $c$.
For a batch of size (B) containing text (T) $\mathbf{x}$ and images (I) $\mathbf{y}$, the contrastive loss is formulated by taking the negative Lorentzian distance as the similarity metric, as follows:
\begin{equation*}
    \mathrm{sim}(f(T), g(I)) = -d_\mathcal{L} (\mathbf{x}, \mathbf{y})
\end{equation*}

%
\noindent\textbf{Reconstruction-guided Contrastive Learning for 3D Point Clouds Embeddings}~ 
The training paradigm of the 3D Point Cloud encoder in this work is based on the reconstruction-guided contrastive learning framework ReCon~\cite{qi2023contrastReCon}. It is consistently observed that contrastive models focus mainly on a global field, in contrast to generative models which exhibit a preference for focused local attention, leading to a task conflict in naive multi-task representation learning settings~\cite{qi2023contrastReCon, xie2023revealing}. 
Following ReCon, we consider the objective as ensemble representation distillation, encouraging the 3D Point Cloud encoder to learn disentangled knowledge representation. 
Both contrastive and generative methods are seen as student-teacher paradigms, unified as ensemble distillation from multiple teachers, where the generative model also acts as a "teacher" guiding the contrastive learning.
Meanwhile, reconstruction guidance enhances the contrastive learning process by improving generalization, stability, and training efficiency.
%
The ReCon loss $\mathcal{L}^{\text{ReCon}}=\mathcal{L}^{\text{Rec}} + \mathcal{L}^{\text{Con}}$ ensembles the cross-modal contrastive learning loss using the positive-only representation learning with Smooth $\ell_1$ loss $\text{Smooth-}\ell_1(\cdot,\cdot)$, and the reconstruction guidance loss, constructed as the masked point modeling reconstruction following~\cite{pang2022masked}.
Specifically, the loss $\mathcal{L}^{\text{Con}}$ can be written as:
\begin{equation}
\begin{aligned}
\mathcal{L}^{\text{Con}} = 
& \sum_{i=1}^{|\mathcal{B}|}\Big[
\text{Smooth-}\ell_1\big(\mathbf{z}, \texttt{stopgrad}(\mathbf{x}\big) + 
\\ 
&\text{Smooth-}\ell_1\big(\mathbf{z}, \texttt{stopgrad}(\mathbf{y}\big)\Big],
\end{aligned}
\end{equation}
where $\texttt{stopgrad}(\cdot)$ is the \textit{stop-gradient} operation, which prevents gradients from back-propagating to the image or text teachers, i.e., keeping the text encoder and image encoder of MERU frozen in this work.
Given the predicted point patches $P_{pre}$ and ground truth $P_{gt}$, we have 
\begin{equation}\label{eq:mpm}
\begin{aligned}
    \mathcal{L}^{\text{Rec}} =
    \frac{1}{|P_{pre}|} 
    & \sum_{P'\in P_{pre}}
    \mathop{\min}\limits_{P\in\mathcal{P}_k} \|P'-P\|^2_2 + 
    \\
    & \frac{1}{|P_{gt}|} 
    \sum_{P\in P_{gt}}    \mathop{\min}\limits_{P'\in\mathcal{P}_{pre}} \|P'-P\|^2_2.
\end{aligned}
\end{equation}

%
%


\section{Approach}
In this section, we first introduce our basic model and then analyze hierarchy relations across text, image, and 3D Point Cloud.
Finally, we present novel hyperbolicity-enhancing pre-training losses that promote preserving hyperbolic modeling capabilities.

\subsection{Overall Architecture and Hierarchy Relations}

\textbf{Basic Model}~ 
When addressing 3D Point Clouds, current hyperbolic contrastive learning methods have not been extended to the 3D Point Clouds modal, ignoring the importance of its underlying hierarchical prior.
Therefore, we extend the reconstruction-guided contrastive learning framework in \cite{qi2023contrastReCon}, aiming to transfer knowledge to the point cloud encoder from a pre-trained hyperbolic language-image model MERU \cite{Ramasinghe2024}, which has learned paired image-text embeddings in the hyperbolic space, facilitate the learning of hierarchical 3D Point Cloud embeddings.
Note that we employ Point-MAE \cite{pang2022masked} as the point cloud encoder. 
Single-modal 3D Point Cloud inputs, and cross-modal inputs including rendered RGB images and text descriptions, are encoded as sequential tokens. 
During contrastive learning, the 3D token embeddings are masked for generative reconstruction, while the mask is disabled during inference. 
The obtained 3D Point Cloud embeddings and global queries are then fed to the decoder, which shares the same architecture as the encoder. The queries are learnable and supervised by contrastive learning.
However, the objective of the 3D Point Cloud encoder in ReCon does not include a hierarchical representation-related loss function or target, meaning that the obtained 3D Point Cloud embeddings do not explicitly capture hierarchical knowledge. 
To address this, we further introduce appropriate regularizers to unify the objective for effective hierarchical 3D representation learning and understanding, while also addressing differences in data patterns and tasks across modalities.


\noindent\textbf{Hierarchy Relation Analysis}~ 
We aim to effectively and explicitly represent the underlying hierarchy structures, whether within a modality (intra-modal) or between different modalities (inter-modal), through positional relationships. 
Specifically, we consider how the relationships between point clouds, text, and images should be structured, focusing on the position of 3D modalities in the hierarchy.

In hyperbolic space, as the distance from the center increases, the available embedding area for features expands exponentially. 
Given that general concepts require less representational space than specific ones, simpler, more common concepts should be mapped closer to the center of the space than more detailed concepts.
Thus,
\begin{itemize}
    \item From the perspective of inter-modal hierarchy relationship, we consider the entailment property used in modeling hierarchical concepts in WordNet~\cite{DBLP:conf/icml/GaneaBH18}, which exhibits the transitive property. This can be applied to construct inter-modal hierarchy relations, meaning if $\mathbf{x}$ entails $\mathbf{y}$ and $\mathbf{y}$ entails $\mathbf{z}$, then $\mathbf{x}$ entails $\mathbf{z}$.

    \item From the perspective of intra-modal, we discuss the alignment of 3D Point Cloud embeddings to Text-Image embeddings, considering the hierarchical \textit{whole $\rightarrow$ part} composition relation inside 3D Point Cloud embeddings and introduce the quantitative analysis of the hierarchical relationships within the same modal.
\end{itemize}

\subsection{Hyperbolic Entailment Regularization}
\label{sec:entail}
To better match the representation of the 3D object's point clouds, existing 3D cross-modal representation learning methods primarily average the image features extracted by single-view 2D foundation models from multi-view images, which are projections of the original 3D object. 
Recently, methods like DETR, BLIP2, and ShapeLLM \cite{qi2025shapellm} have further proposed to adaptively select and distill views to describe the 3D object better.
However, while images can supplement information about material and color that 3D Point Clouds might not provide, they can only serve as approximations or augmentations for accurately representing the geometric structure of the 3D object.
Therefore, to perform cross-modal learning better, we must acknowledge the differences between image embeddings and 3D Point Cloud embeddings and capture their relationship, which enhances comprehensive 3D shape information understanding.

In a nutshell, multi-view image embeddings are entailed within the hyperbolic cones of the 3D Point Cloud embeddings, and 3D Point Cloud embeddings are entailed within the hyperbolic cones of the text embeddings. 
This builds a hierarchy structure of the cross-modal feature space, establishing a partial order among Text-Image-point cloud pairs.

We extend the entailment loss~\cite{DBLP:conf/icml/DesaiNR0V23}, generalizing to capture the visual-semantic hierarchy across Text-Image-point cloud three modalities.
Using the half-aperture of a text embedding and the exterior angle between a Text and Image embedding as examples, as
\begin{equation*}
    \text{aper}(\mathbf{x}) = \sin^{-1} \left( \frac{2K}{\sqrt{c} \; \lVert \mathbf{x}_{space} \rVert} \right), \; \lVert \mathbf{x}_{space} \rVert \ge \frac{2K}{\sqrt{c}},
\end{equation*}
where $\frac{2K}{\sqrt{c} \; \lVert \mathbf{x}_{space} \rVert}$ will be clamped to $1 - \epsilon$, where $ \epsilon = 10^{-8}$ for training stability \cite{DBLP:conf/icml/DesaiNR0V23}.
The exterior angle $ \text{ext}(\mathbf{x}, \mathbf{y}) = \pi - \angle \mathbf{O} \mathbf{x} \mathbf{y}$ given by the origin $\mathbf{O}$, $\mathbf{x}$, and $\mathbf{y}$ is then
\begin{equation*}
\begin{aligned}
    \def\cvl{c \; \langle \mathbf{x},\mathbf{y} \rangle _\mathcal{L}}
    \text{ext}(\mathbf{x}, \mathbf{y}) = \cos^{-1} \left( \frac{y_{time} + x_{time} \; \cvl{}}{\lVert \mathbf{x}_{space} \rVert \sqrt{\left( \cvl \right)^2 - 1}} \right)
\end{aligned}
\end{equation*}
Finally, the entailment loss $mathcal{L}_{\text{entail}}$ is formally written as: 
\begin{equation*}
\begin{aligned}
    \mathcal{L}_{\text{entail}}(\mathbf{x}, \mathbf{y},  \mathbf{z}) = \max(0, \; & \text{ext}(\mathbf{x}, \mathbf{y}) + \text{ext}(\mathbf{y}, \mathbf{z})- \\ & \text{aper}(\mathbf{x})- \text{aper}(\mathbf{y})).
\end{aligned}
\end{equation*}
The entailment loss forces all image embeddings to match the point cloud embeddings within the cones that emanate from the point cloud embeddings and all point cloud embeddings to match the text embeddings within the cones that emanate from the text embeddings. 
Images still adhere to the principle that a corresponding text embedding represents a more abstract concept, and we build the inter-modal hierarchy between the embeddings of the new modality 3D Point Cloud and the existing Text-Image embeddings.

To encourage better distribution of embeddings on the hyperbolic space, we follow an objective that maintains a controllable or adaptive modality gap and further constructs the semantic hierarchy structure across text, 2D images, and 3D Point Clouds \cite{Ramasinghe2024}. 
Specifically, we ensure that the centroid of text embeddings is closer to the origin than the centroid of visual embeddings, and the centroid of 3D Point Cloud embeddings should be closer to the origin than the centroid of the 2D image embeddings.
Obtaining the centroid of a set of points in the hyperbolic space $\mathbb{H}$ is not as straightforward as the Euclidean setting. 
This is called the Einstein midpoint, and it is easier to obtain via converting to Klein coordinates $\mathbb{K}$.
A point on the hyperboloid model can be converted to Klein coordinates $\mathbf{k}$ and back via projections, and the the centroid takes the following form:
\begin{equation*}
\begin{aligned}
    \text{Centroid}_{\mathbb{H}}\left(\mathbf{x}\right) = \Pi_{\mathbb{K}\to\mathbb{H}}\left(\frac{\sum_{j=1}^{N}{\gamma_j\Pi_{\mathbb{H}\to\mathbb{K}}(\mathbf{x}_j)}}{\sum_{j=1}^{N}{\gamma_j}}\right), \\
\mathbf{x} = \left\{\mathbf{x}_j\right\}_{j=1}^{N}, \gamma_j=\frac{1}{\sqrt{1-c\|\Pi_{\mathbb{H}\rightarrow\mathbb{K}}(\mathbf{x}_j)\|^2}}
\end{aligned}
\end{equation*}
where $\gamma_j$ denotes the Lorentz factors.
Formally, let $\mathbf{X}_e$, $\mathbf{Y}_e$, and $\mathbf{Z}_e$ be the Einstein midpoint of a set of text, 3D Point Cloud embeddings and image embeddings, respectively. 
Then, our regularization takes the following form:
\begin{equation*}
\begin{aligned}
\centering
p & = d_\mathcal{L} (\mathbf{O}, \mathbf{p}) = \sqrt{\sfrac{1}{c}} \cdot \cosh^{-1} ( -c \; \langle \mathbf{O},\mathbf{p} \rangle_\mathbb{L}) \\
& =\sqrt{\sfrac{1}{c}} \cdot \cosh^{-1} (\sqrt{\frac{1}{c}+\lVert \mathbf{p}_{space} \rVert^2}), \\
q  & = d_\mathcal{L} (\mathbf{O}, \mathbf{q}),
r  = d_\mathcal{L} (\mathbf{O}, \mathbf{r}), \\
\mathcal{L}_{\text{cent}} & = \lVert \mathbf{Z}_e - p \rVert^2 + \lVert \mathbf{Y}_e - q \rVert^2 + \lVert \mathbf{X}_e - r \rVert^2
\end{aligned}
\end{equation*}
where  is the Euclidean norm and $p > q > r > 0$ to ensure that
the centroid relationships, avoiding the diversity collapse of visual embeddings.


\subsection{Alignment Loss and Hyperbolicity Analysis}
\label{sec:alignment}
In addition to achieving the alignment of 3D Point Cloud embeddings to Text-Image embeddings, we refine the internal hierarchical structure of the point cloud embeddings by considering the hierarchical \textit{whole $\rightarrow$ part} composition relation.
Specifically, during training, the 3D Point Cloud encoder Point-MAE infers twice: once with full-size point clouds as input and once with a masked point cloud, resulting in a part embedding. This process helps capture the hierarchical relationships within the point cloud modal.

Next, we introduce a quantitative analysis method to analyze the hierarchical structure of the obtained point cloud embeddings.
Gromov $\delta$-hyperbolicity is a geometric metric that quantifies the deviation of a given metric space from an exact tree metric \cite{gromov1987hyperbolic}.
The simplest discrete metric space possessing hyperbolic properties is a tree (in the sense of graph theory) endowed with the natural shortest path distance. 
A lower $\delta$-hyperbolicity value, or equivalently, a higher degree of hyperbolicity, indicates a more tree-like structure within the space, and $\delta=0$ for trees. 

We follow the efficient way to compute $\delta$ presented in~\cite{fournier2015computing} and applied in \cite{pavlovna2024development, DBLP:conf/cvpr/KhrulkovMUOL20}. Specifically, we sample $N$ samples and find $\delta_{rel}$ based on the distance matrix of point sets.
Since we compute the hyperbolicity values of features that embedded continuous space, i.e., Lorentz Space, the distance matrix is based on their geodesic distances.
Moreover, $\delta_{rel}$ is a scale-invariant metric defined as: 
\begin{equation}
\label{eq:hyperbolicity}
  \delta_{rel}(X) = \frac{2\delta(X)}{\mathrm{diam}(X)},  
\end{equation}
where $\mathrm{diam}(X)$ denotes the set diameter (maximal pairwise distance). 
By construction, $\delta_{rel}(X) \in [0, 1]$ specifies how close embeddings are set to a hyperbolic space.


\subsection{Contention Between Modal and Tasks}

For the setting of visual scene understanding in computer vision, the down-streamed models must understand both the geometry and semantics of the scene by dealing with our learned visual embeddings simultaneously, leading to a multi-task representation learning problem.
Considering that the different regularizers are independently propagated, the coverage speed of intra-modal learning and inter-modal learning was observed to be different during our pre-training \cite{huang2023clip2point}.
Prior approaches to simultaneously learning multiple tasks use a naive weighted sum of losses, where the loss weights are uniform or manually tuned. Performance is highly dependent on an appropriate choice of weighting between each task’s loss.
Therefore, to ensure the effective use of the regularizers proposed in the above sections, this work automatically weighs multiple loss functions based on each task's homoscedastic uncertainty \cite{kendall2018multi}.
All objectives can be seen as regression-based reconstruction tasks \cite{huang2024zeroshape}. 
Thus, we adopt this multi-task loss to balance these terms.
The overall joint loss function is formulated as
\begin{equation*}
\mathcal{L} = \sum_i (e^{-s}\mathcal{L}_i + s), 
\mathcal{L}_i\in \{\mathcal{L}_{\text{cent}}, \mathcal{L}_{\text{entail}}, \mathcal{L}_{\text{Rec}}, \mathcal{L}_{\text{Con}}\}
\end{equation*}

\section{Experimental Results}
In this work, we apply the reconstruction-guided contrastive learning strategy~\cite{qi2023contrastReCon} to train a 3D Point Cloud encoder, building on the hyperbolic Text-Image model MERU~\cite{DBLP:conf/icml/DesaiNR0V23} and extending its multi-modal capabilities to obtain hierarchical 3D Point Cloud embeddings.
Consequently, in this section, we mainly conducted experiments to evaluate the 3D Point Cloud encoder to answer the following two pivotal Research Questions (RQs):
\textbf{RQ1}: What advantages does our method offer in achieving hierarchical 3D Point Cloud embeddings? \textbf{RQ2}: What improvements do our hierarchical 3D Point Cloud embeddings bring to the 3D Point Cloud downstream tasks?
To answer \textbf{RQ1}, we analyze the hierarchical relationships among embeddings across language and vision modalities in Section~\ref{sec:an_to_rq1}.
To answer \textbf{RQ2}, we compare our method with state-of-the-art 3D Point Cloud methods on a variety of datasets and tasks in Section~\ref{sec:an_to_rq2}, demonstrating the improvements brought by our hierarchical 3D point cloud embeddings.

\noindent\textbf{Implementation Details}~
All the experiments are conducted on a GeForce RTX 4090 24 GB for all the experiments.
We use CLIP and MERU as our teacher models: the CLIP-based approach uses the default CLIP-ViT-B/16, and the MERU-based method uses the base MERU model with ViT-B/16 as the image encoder and the same text encoder as CLIP \cite{DBLP:conf/nips/VaswaniSPUJGKP17}.
%
We maintain consistent settings of the total train epoch, learning rate schedule, and optimizer configuration with maximum learning rate $5e-4$, AdamW optimizer, cosine learning rate schedule with $10$ warm-up steps, and batch size $128$. 
To ensure curvature parameter $c$ and text/image/point clouds embedding scales $\alpha_{txt}/\alpha_{img}/\alpha_{pts}$ stay positive, these scalar hyperparameters are parameterized on the logarithmic scale and are consistent with the MERU model: $c$ is initialized with pre-trained MERU checkpoint and fixed.
To maintain stability and avoid numerical issues, we employ trigonometric functions with values clamped to the range $[1e-8, 1-1e-8]$.

\begin{table}[b]
\caption{Average Hyperbolicity.}
\label{table:1}
\centering
\resizebox{0.45\textwidth}{!}{
\begin{tabular}{cccc}
\toprule
 Methods & Text & Image & Point Clouds\\ 
\midrule
CLIP-based &
  \begin{tabular}[c]{@{}c@{}}0.2695 $\pm$ 0.02287\\(0.4375$\pm$0.0199) \end{tabular} &
  \begin{tabular}[c]{@{}c@{}}0.3116$\pm$0.014 \\ (0.3075$\pm$0.0203)\end{tabular} &
  \begin{tabular}[c]{@{}c@{}}0.3639\\ $\pm$ 0.0216\end{tabular} \\ \cline{1-4} 
MERU-based &
  \multirow{3}{*}{\begin{tabular}[c]{@{}c@{}}0.2331$\pm$0.0154\\(0.3745$\pm$0.0121)\end{tabular}} &
  \multirow{3}{*}{\begin{tabular}[c]{@{}c@{}}0.3535$\pm$0.0350\\(0.3241$\pm$0.0284)\end{tabular}} &
  \begin{tabular}[c]{@{}c@{}}0.3288 \\ $\pm$ 0.03014 \end{tabular} \\ \cline{1-1}\cline{4-4} 
MERU modified-based &      &       & \begin{tabular}[c]{@{}c@{}}0.1716\\ $\pm$ 0.02735\end{tabular} \\ 
\bottomrule
\end{tabular}
}
\end{table}

\begin{figure}[h]
\centering
\begin{subfigure}[t]{.23\textwidth}
\centering
    \includegraphics[width=39mm, height=28mm]{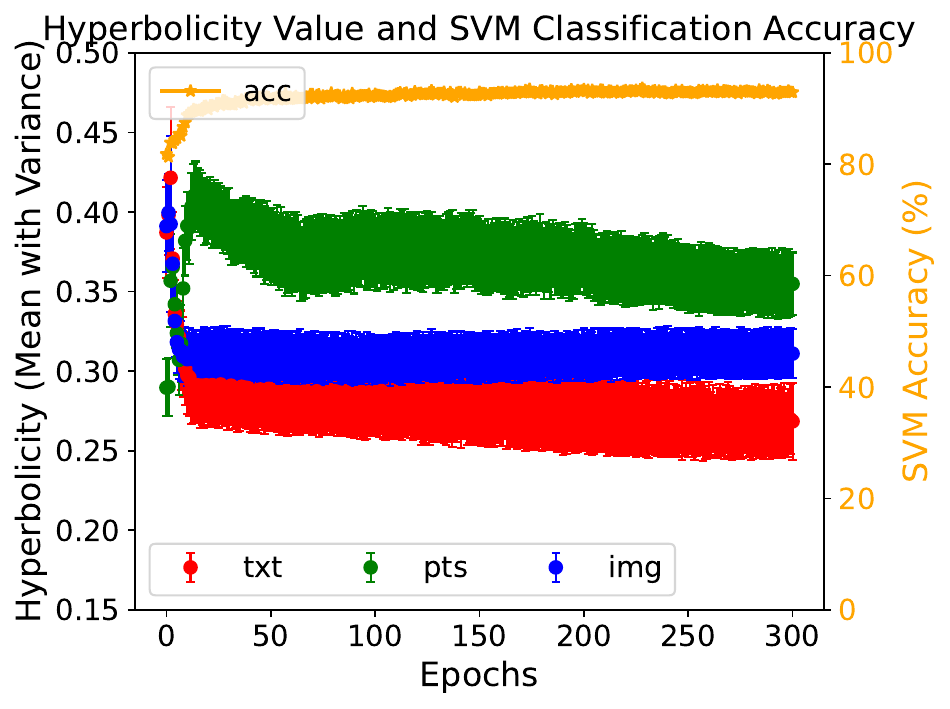}
    \caption{Curvature Curves (CLIP-based)}
    \label{fig:hyperbolicity-a}
\end{subfigure}
\begin{subfigure}[t]{.24\textwidth}
\centering
    \includegraphics[width=39mm, height=28mm]{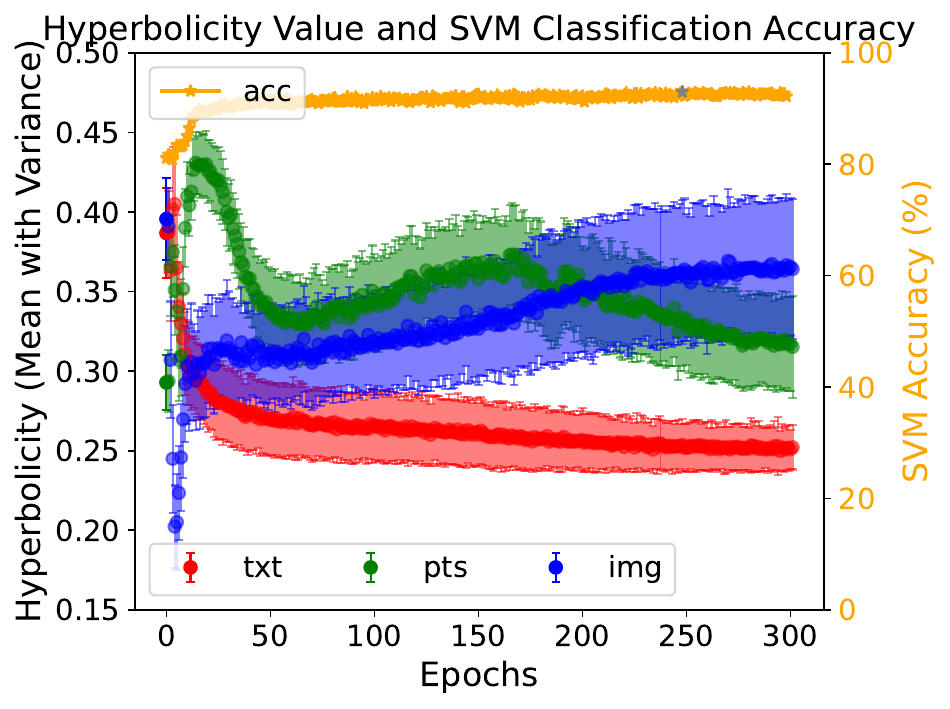}

    \caption{Curvature Curves (MERU-based)}
    \label{fig:hyperbolicity-a2}
\end{subfigure}
\hfill
\begin{subfigure}[t]{.5\textwidth}
    \centering   
    \includegraphics[width=40mm, height=28mm]{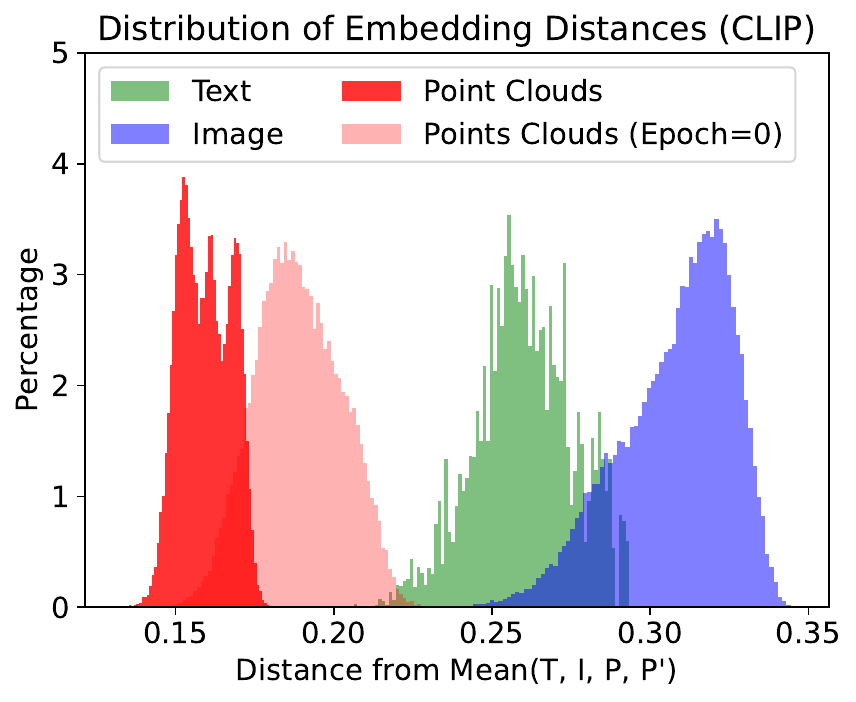}
    \includegraphics[width=40mm, height=28mm]{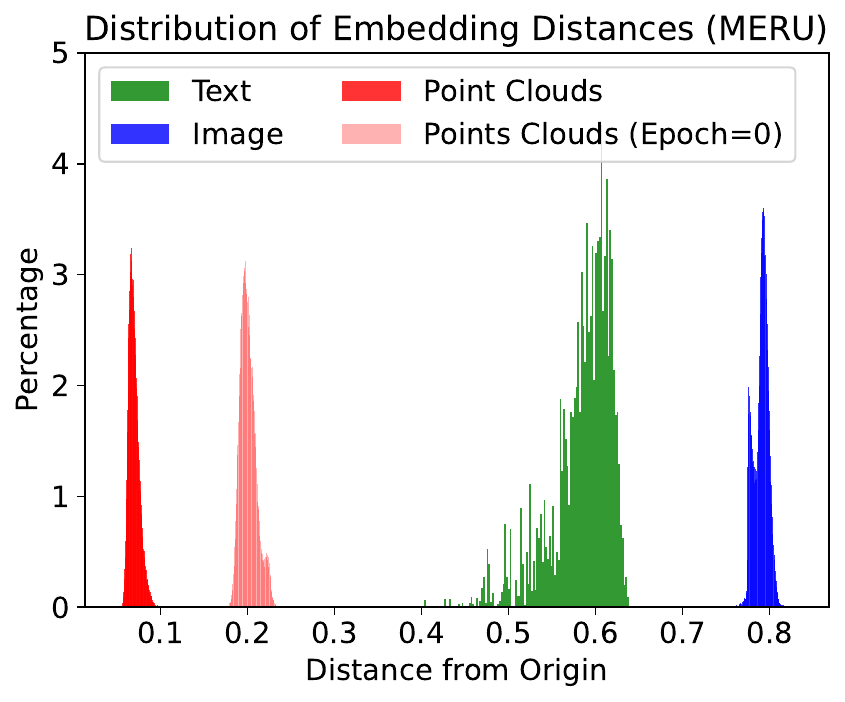}
    \caption{Distribution of Embedding Distance}
    \label{fig:hyperbolicity-b}
\end{subfigure}
\caption{Hyperbolicity coverage curves and distribution of embedding distances of text embeddings, image embeddings, and point cloud embeddings.
}
\label{fig:hyperbolicity}
\end{figure}

\subsection{Hierarchical Embedding Analysis}
\label{sec:an_to_rq1}
We calculate the $\delta$-hyperbolicity values according to Equation~\ref{eq:hyperbolicity} and present in Table \ref{table:1} and Figure~\ref{fig:hyperbolicity}.
We measure the hyperbolicity of embeddings within individual randomly shuffled batches, each containing $128$ samples. 

The final $\delta$ result is the average and standard deviation across all $409$ batches.
The images are projected from the 3D Point Clouds. 
The 3D Point Cloud models are from ShapeNet \cite{wu20153d}, which are organized into WordNet synsets naturally with a hierarchical structure.

\begin{figure}[t]
\centering
\begin{subfigure}[t]{.5\textwidth}
    \centering
\includegraphics[width=40mm, height=35mm]{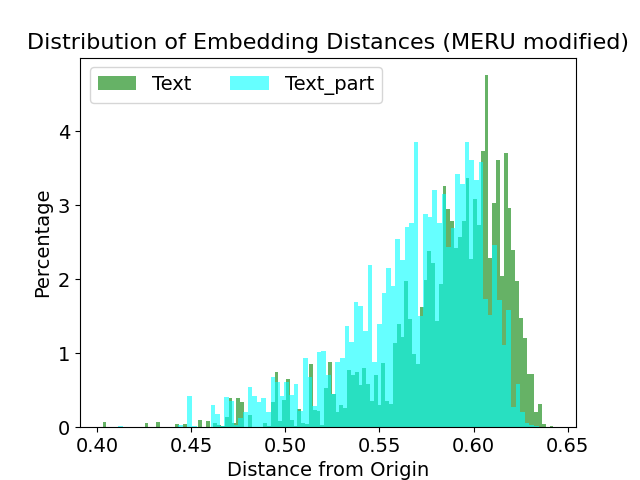}
\includegraphics[width=40mm, height=35mm]{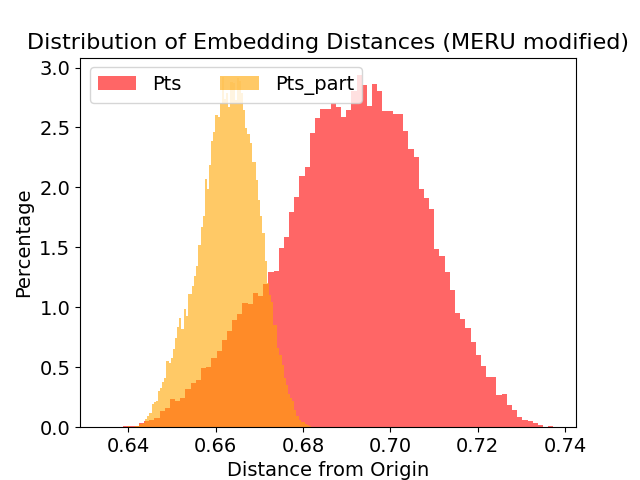}
    \caption{Distribution of embedding distances between text and 3D point cloud embeddings shows the \textit{whole $\rightarrow$ part} composition relation.}
    \label{fig:distance-a}
\end{subfigure}
\begin{subfigure}[t]{.5\textwidth}
    \centering
\includegraphics[width=60mm, height=45mm]{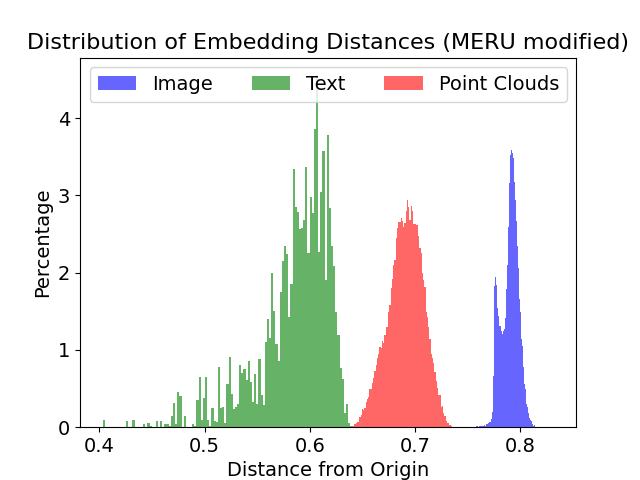}
    \caption{Distribution of embedding distances between text, image, and 3D point cloud embeddings demonstrates that the inter-modal hierarchical relationship.}
    \label{fig:distance-b}
\end{subfigure}
\caption{
Analysis of Embedding Distances for Text, Image, and Point Cloud Data via our approach (MERU (modified)).
}
\label{fig:distance-distribution}
\end{figure} 

\begin{figure}[h]
\centering
\begin{subfigure}[t]{.5\textwidth}
    \centering
\includegraphics[width=40mm, height=30mm]{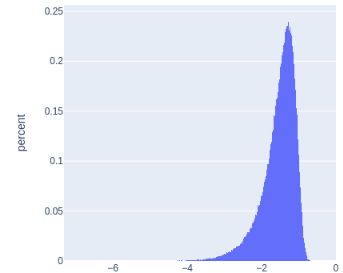}
\includegraphics[width=40mm, height=30mm]{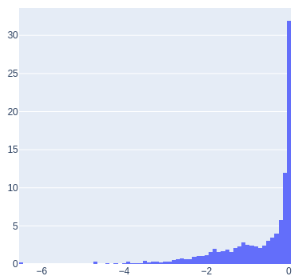}
    \caption{Log-scale distribution of cosine similarities in the dictionary atoms and frequency of latent features, with the y-axis representing the percentage.}
    \label{fig:distance-a}
\end{subfigure}
\\
\begin{subfigure}[t]{.5\textwidth}
    \centering
\includegraphics[width=80mm, height=30mm]{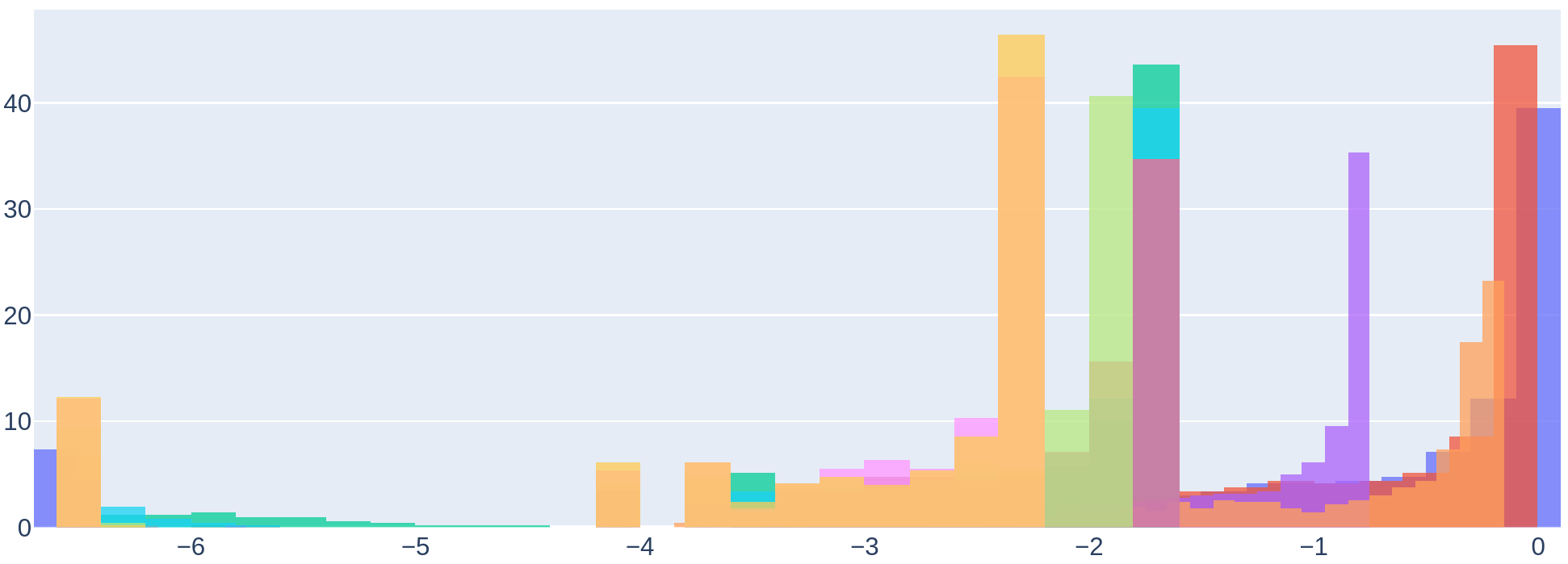}
    \caption{Distribution of sparse features of 10 different classes randomly sampled from ShapeNet-55 dataset (with different colors to represent each class).}
    \label{fig:distance-aa}
\end{subfigure}
\caption{
Disentangled analysis for our obtained 3D Point Cloud embeddings by dictionary learning approach.
}
\label{fig:dic_ana}
\end{figure}

\begin{table*}[h]
\centering
\caption{Part segmentation results on the ShapeNetPart dataset.
}
\label{tab: partseg}
\resizebox{\textwidth}{!}{
\begin{tabular}{l|cc|cccccccccccccccc}
\toprule
Methods        & mIoU$_C$       & mIoU$_I$      & aero & bag  & cap  & car  & chair & earphone & guitar & knife & lamp & laptop & motor & mug  & pistol & rocket & skateboard & table \\ \midrule
PointNet \cite{qi2017pointnet}       & $80.39$          & $83.7$          & $83.4$ & $78.7$ & $82.5$ & $74.9$ & $89.6$  & $73.0$     & $91.5$   & $85.9$  & $80.8$ & $95.3$   & $65.2$  & $93.0$ & $81.2$   & $57.9$   & $72.8$       & $80.6$  \\
PointNet++ \cite{qi2017pointnet++}     & $81.85$          & $85.1$          & $82.4$ & $79.0$ & $87.7$ & $77.3$ & $90.8$  & $71.8$     & $91.0$   & $85.9$  & $83.7$ & $95.3$   & $71.6$  & $94.1$ & $81.3$   & $58.7$   & $76.4$       & $82.6$  \\
DGCNN \cite{wang2019dynamic}          & $82.33$          & $85.2$          & $84.0$ & $83.4$ & $86.7$ & $77.8$ & $90.6$  & $74.7$     & $91.2$   & $87.5$  & $82.8$ & $95.7$   & $66.3$  & $94.9$ & $81.1$   & $63.5$   & $74.5$       & $82.6$  \\ 
Transformer \cite{yu2021point}    & $83.42$          & $85.1$          & $82.9$ & $85.4$ & $87.7$ & $78.8$ & $90.5$  & $80.8$     & $91.1$   & $87.7$  & $85.3$ & $95.6$   & $73.9$  & $9$$4.9$ & $83.5$   & $61.2$   & $74.9$       & $80.6$  \\
Point-BERT \cite{yu2021point}     & $84.11$          & $85.6$          & $8$$4.3$ & $84.8$ & $88.0$ & $79.8$ & $91.0$  & $81.7$     & $91.6$   & $87.9$  & $85.2$ & $95.6$   & $75.6$  & $94.7$ & $84.3$   & $63.4$   & $76.3$       & $81.5$  \\
Point-GT-G \cite{li2022geodesic} & $83.94$          & $85.9$          & $84.7$ & $83.7$ & $89.4$ & $80.4$ & $91.2$  & $77.0$     & $91.7$   & $87.6$  & $85.6$ & $96.0$   & $74.0$  & $95.3$ & $84.6$   & $62.7$   & $77.5$       & $81.7$  \\
Point-GT-DM \cite{li2022geodesic} & $84.15$ & $85.8$ & $84.3$ & $84.5$ & $88.3$ & $80.9$ & $91.4$  & $78.1$     & $92.1$   & $88.5$  & $85.3$ & $95.9$   & $77.1$  & $95.1$ & $84.7$   & $63.3$   & $75.6$       & $81.4$  \\ 
\midrule
CLIP-based & 84.49 & 86.14 & 85.17 & 82.63 & 89.28 & 81.11 & 91.62 & 77.40 & 92.26 & 88.70 & 86.16 & 95.96 & 77.31 & 95.54 & 83.99 & 62.66 & 78.75 & 81.55
\\
MERU-based & 84.70 & 86.28 & 85.42 & 84.47 & 89.06 & 81.22 & 91.60 & 75.15 & 92.01 & 88.40 & 86.23 & 96.16 & 77.40 & 95.02 & 84.96 & 65.61 & 77.05 & 81.96\\
MERU (modified)-based & 84.91 &86.49 & 85.51& 85.53&88.88&81.02&91.42&80.39&91.47&88.83&86.86&97.17&77.08&95.92&84.15&62.46&77.38&82.72\\
\bottomrule
\end{tabular}
}
\end{table*}

\begin{table}[h]
\caption{Classification performance of models fine-tuned on ModelNet40 and ModelNet10.}
\label{table:finetuned}
\centering
\resizebox{0.46\textwidth}{!}{
\begin{tabular}{ccc} 
\toprule
Methods & ModelNet40 1k(8k) & ModelNet10 \\
\midrule
CLIP-based & 93.44(94.12) & 95.04 \\
MERU-based & 93.0713(93.64) & 94.71\\
MERU (modified)-based & 93.64(94.30) & 95.37\\
 \bottomrule
\end{tabular}
}
\end{table}

\begin{table}[h]
\caption{Few-shot performance on ModelNet40.}
\label{table:few-shot}
\centering
\resizebox{0.48\textwidth}{!}{
\begin{tabular}{ccccc}
\toprule
Methods & \multicolumn{2}{c}{5-way} & \multicolumn{2}{c}{10-way} \\ \cline{2-5} 
& 10-shot     & 20-shot     & 10-shot      & 20-shot     \\
\midrule
ReCon      & 97.3        & 98.9        & 93.3         & 95.8        \\
CLIP-based & 97.4        & 98.7        & 94.0         & 95.95       \\
MERU-based & 97.1  &  99.1 & 94.1 & 95.6  \\
MERU (modified)--based & 97.4  &  99.1 & 93.5 & 95.9  \\
\bottomrule
\end{tabular}
}
\end{table}

Figure~\ref{fig:hyperbolicity-a} and \ref{fig:hyperbolicity-a2} show that text embeddings coverage to a lower degree of hyperbolicity compared to vision embeddings.
Nevertheless, both CLIP and MERU-based models learned embeddings that align with a hyperbolic structure across text and vision modalities, similar to the findings of \cite{yang2024enhancing} and HyCoRe \cite{montanaro2022rethinking} that hierarchies emerge naturally without any regularizers forcing them.
This suggests that embeddings of the text and vision modalities are highly organized and exhibit non-Euclidean hyperbolic patterns.
Figure~\ref{fig:hyperbolicity-a2} indicates that the reconstruction-guided approach appears to be effective. Compared to CLIP-based methods, MERU-based methods lead to point cloud embeddings converging to lower hyperbolicity values in the latter stages of training. This suggests a stronger hierarchical structure in the inter-modal 3D point cloud embeddings.
Table \ref{table:1} shows the hyperbolicity of the learned embeddings in the first row of the grid and the target embeddings in the second row, across all three modalities. We observe a distinct difference in hyperbolicity between the learned embeddings and the target embeddings obtained from the teacher models (CLIP or MERU), especially when there is a direct hyperbolicity regularizer for the embeddings.
Meanwhile, regardless of whether the teacher model is CLIP-based or MERU-based, the obtained text embeddings exhibit lower hyperbolicity compared to the target embeddings, while image embeddings show the opposite.

Further, we demonstrate the embedding space structures by plotting the distances of all training data embeddings from [ROOT].
Following MERU, for hyperbolic-geometry-based embedding space, the [ROOT] is fixed as the origin point of the Lorentz hyperboloid, which encompasses the entire representation space and is the [ROOT] of all embeddings.
The [ROOT] position of CLIP is estimated as the embedding vector that has the minimum distance to all embeddings in the training dataset. Therefore, the average of all modality embeddings is taken, followed by L2 normalization.
As illustrated in Figure~\ref{fig:hyperbolicity-b}, the text embeddings and image embeddings remain overlapped for CLIP, while exists a gap in MERU, which is more consistent with the inherent modality gap \cite{Ramasinghe2024}.
Meanwhile, the obtained point cloud embeddings are driven to the origin $\mathbf{O}$ and close to $\mathbf{O}$ than text embeddings, which in turn are closer than the image embeddings.
However, according to the hierarchy relationship, the 3D Point Cloud embeddings should be entailed in the text modality, indicating farther $\mathbf{O}$ than text embeddings.
%

Moreover, we explore the effectiveness of the external regularizers forcing embeddings to have explicit textual entailment relationships.
As demonstrated in Figure~\ref{fig:distance-a}, the distribution of embedding distances between text and 3D point cloud embeddings (Pts) reveals the \textit{whole $\rightarrow$ part} composition relation. 
Specifically, the embeddings of `part' concepts (Text and Pts) tend to be more common and require less embedding space, placing them closer to the origin. 
This suggests that `part' concepts are more frequently encountered and thus have more compact embeddings, 
while `whole' concepts (Text\_ part and Pts\_ part) are more specific and occupy a larger and more diverse embedding space, resulting in greater distances from the origin. 
Figure~\ref{fig:distance-b} shows that our regularizers explicitly construct hierarchical relations across intra-modal embeddings while maintaining their modality gap.
For the intra-modal hierarchical relationship analysis, we have observed that with the MERU-based method, the position relation in Figure \ref{fig:hyperbolicity-b} does not fully correspond to our entailment and centroid conditions. Specifically, the centroid of 3D point cloud embeddings is not positioned between the centroids of image and text embeddings as expected. 
Figure \ref{fig:distance-b} shows that our regularizers, named our MERU (modified)-based method explicitly construct hierarchical relations across intra-modal embeddings while maintaining their modality gap, better aligning with the inherent modality differences.

\subsection{Dictionary Learning Analysis}
We trained a Sparse Autoencoder (SAE)~\cite{Rajamanoharan2024GatedSAE}, to obtain sparse codes of 3D Point Clouds features, while the decoder is a linear layer whose weights act as a traditional dictionary. 
Each column in this dictionary represents an atom. 
We set the dimension of each atom to $512$ and the total number of atoms to five times the feature dimension, totaling $2560$ atoms.
Figure \ref{fig:dic_ana} shows the logarithmic frequency of sparse codes and the cosine similarities between all atoms, indicating the sparsity of the data structure. 
The atoms also exhibit low similarity, suggesting that our dictionary has learned highly independent base features. 
Notably, approximately $30\%$ of the atoms show significant activation, as they are frequently activated when encoding the 3D Point Cloud samples.
Furthermore, for samples of different classes, frequently activated atoms display distinct activation levels, indicating that the learned dictionary is discriminative. This also implies that the learned features can be further disentangled and semantically decomposed, which will be explored in future work.

\subsection{Point Cloud Encoder Experiments}
\label{sec:an_to_rq2}
We conduct the evaluation on tasks including fine-tuning classification, few-shot learning, and part segmentation experiments.
We conduct classification, and few-shot learning experiments on ModelNet40 and ModelNet10 datasets \cite{wu20153d}, which are all popular 3D object datasets. 
We use data augmentation operations during training, following ReCon \cite{qi2023contrastReCon}, including standard random scaling and translation.


\subsubsection{Fine-tuned Classification Results.}
We finetune on three model variants on ModelNet40, and ModelNet10.
On ModelNet40, we report two different settings with $1,024$ (1k) points and $8,192$ (8k) points respectively.
Since the real-world dataset will inevitably be affected by noise or occlusions, it is a much more challenging dataset for point cloud analysis methods. 
The results are shown in Table~\ref{table:finetuned}.
In the comparison, our MERU (modified) significantly boosts the performance without voting, achieving higher accuracy scores than the other three baselines including ReCon, CLIP-based, and MERU-based approaches.


\subsubsection{Few-shot 3D object classification}
We conduct few-shot learning experiments on ModelNet40 and zero-shot experiments on ModelNet40 and ModelNet10.
In few-shot experiments, we use the 'K-way N-shot' setting, randomly selecting $K$ classes and $N$ instances per class to form a support set for training. 
We then evaluate the model using other N instances from the same K classes, referred to as the query set.
The results with the setting of \( K \in \{5, 10\} \) and \( N \in \{10, 20\} \) are presented in Table~\ref{table:few-shot}. 
Note that additionally, we did not modify the task head to incorporate hyperbolic operations such as Möbius addition; instead, we continued using the original linear layers and activation functions. The necessity of such modifications can be explored in future work. Nonetheless, Table~\ref{table:few-shot} demonstrates that the proposed MERU (modified)-based approach achieves competitive performance.


\subsubsection{Part Segmentation}
Object part segmentation is a challenging task with a high requirement of model representation capability. 
Table~\ref{tab: partseg} reports the part segmentation results on the ShapeNetPart \cite{yi2016scalable} dataset, providing the best average class mIOU $mIoU_C$ (\%), the best average inctance mIOU, the $mIoU_I$ (\%), as well as the IoU (\%) for each categories.
Recall that Table~\ref{table:1} shows that MERU-based 3D Point Cloud embeddings have lower hyperbolicity than CLIP-based embeddings.
However, MERU-based embeddings are close to the origin, which violates the priors introduced in \cite{montanaro2023towards} that the embeddings of 3D models' components or parts should ideally sit at the border of hyperbolic geometry, where there is exponentially greater capacity.
We suggest that the semantic hierarchy relationships of point cloud embeddings from existing pre-trained hyperbolic Text-Image models also benefit the performance of 3D Point Cloud segmentation tasks, even at the category level, and are important.

\section{Conclusion}
In this work, we extended the application of hyperbolic geometry to multi-modal data, integrating 3D Point Cloud data with Text and Image modalities. 
We further propose hierarchy-enhancing regularizers to align 3D Point Cloud embeddings with hyperbolic Text-Image embeddings and effectively capture both \textit{intra-modal} and \textit{inter-modal} hierarchical knowledge. By establishing a partial order among Text-Image-3D Point Cloud pairs, we constructed hierarchical semantically similar relationships across different modalities, enhancing the interpretability of these embeddings. 
Experimental results demonstrate significant improvements in various point cloud tasks, the effectiveness of our approach in multi-modal embeddings and hierarchical 3D Point Cloud embeddings learning.

\bibliographystyle{ieee_fullname}
\bibliography{main}

\end{document}